\documentclass[letterpaper]{article} 
\usepackage{aaai24}  
\usepackage{times}  
\usepackage{helvet}  
\usepackage{courier}  
\usepackage[hyphens]{url}  
\usepackage{graphicx} 
\usepackage{natbib}  
\usepackage{amsmath} 
\usepackage{bbding}
\usepackage{booktabs}
\usepackage{amssymb}
\usepackage[T1]{fontenc}
\usepackage{caption} 
\usepackage{algorithm}
\usepackage{algorithmic}

\usepackage{newfloat}
\usepackage{listings}
\DeclareCaptionStyle{ruled}{labelfont=normalfont,labelsep=colon,strut=off} 
\floatstyle{ruled}
\newfloat{listing}{tb}{lst}{}
\floatname{listing}{Listing}
\title{Frequency-Controlled Diffusion Model for Versatile Text-Guided Image-to-Image Translation}
\author{
    Xiang Gao,
    Zhengbo Xu,
    Junhan Zhao,
    Jiaying Liu\thanks{Corresponding author. \textbf{Accepted by AAAI 2024}.}
}
\affiliations{
    Wangxuan Institute of Computer Technology, Peking University, Beijing, China \\

    \{gaoxiang1102, icey.x, liujiaying\}@pku.edu.cn
}

\begin{document}

\maketitle

\begin{abstract}
Recently, large-scale text-to-image (T2I) diffusion models have emerged as a powerful tool for image-to-image translation (I2I), allowing open-domain image translation via user-provided text prompts. This paper proposes frequency-controlled diffusion model (FCDiffusion), an end-to-end diffusion-based framework that contributes a novel solution to text-guided I2I from a frequency-domain perspective. At the heart of our framework is a feature-space frequency-domain filtering module based on Discrete Cosine Transform, which filters the latent features of the source image in the DCT domain, yielding filtered image features bearing different DCT spectral bands as different control signals to the pre-trained Latent Diffusion Model. We reveal that control signals of different DCT spectral bands bridge the source image and the T2I generated image in different correlations (e.g., style, structure, layout, contour, etc.), and thus enable versatile I2I applications emphasizing different I2I correlations, including style-guided content creation, image semantic manipulation, image scene translation, and image style translation. Different from related approaches, FCDiffusion establishes a unified text-guided I2I framework suitable for diverse image translation tasks simply by switching among different frequency control branches at inference time. The effectiveness and superiority of our method for text-guided I2I are demonstrated with extensive experiments both qualitatively and quantitatively. Our project is publicly available at: https://xianggao1102.github.io/FCDiffusion/.
\end{abstract}

\begin{figure}[t]
    \centering
    \includegraphics[width=3.3in]{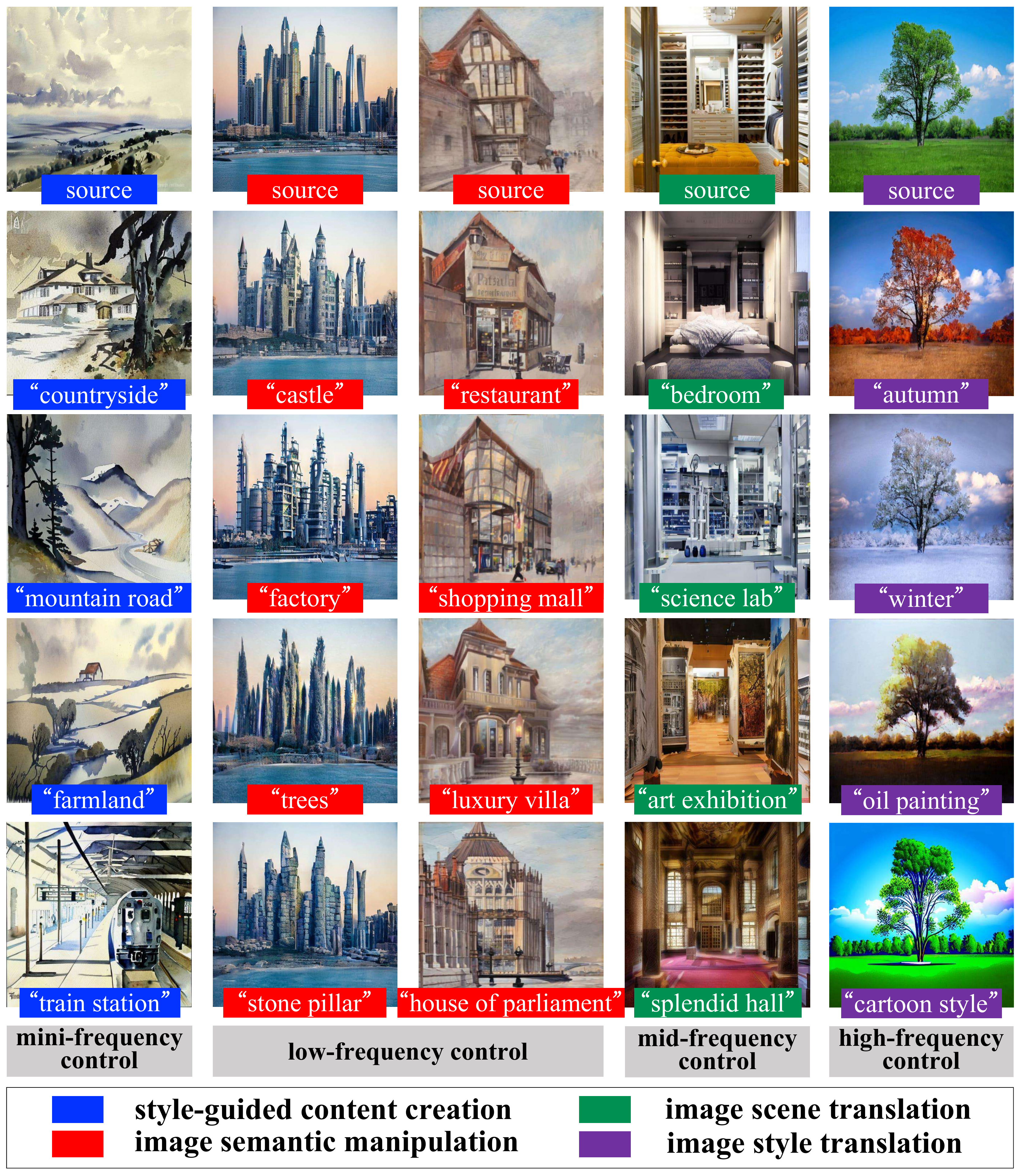}
    \caption{Our FCDiffusion adapts Stable Diffusion to versatile text-guided I2I applications via different types of frequency control, e.g., style-guided content creation realized by mini-frequency control, image semantic manipulation realized by low-frequency control, image scene translation realized by mid-frequency control, and image style translation realized by high-frequency control. \textbf{Better viewed with zoom-in}.}
    \label{fig:cover}
\end{figure}

\section{Introduction}
\label{intro}
Image-to-image translation (I2I) is an attractive computer vision problem. Early I2I methods learn a cross-domain I2I mapping via GANs \cite{goodfellow2014generative}. Given paired training data of two domains, Pix2Pix \cite{isola2017image} establishes a general conditional GAN framework for supervised I2I, derived from which more complex architectures are designed for specific tasks and applications \cite{jiang2019scfont, yi2019apdrawinggan}. Since collecting paired training data is costly or even practically infeasible, unsupervised I2I (UI2I) methods gain rapid popularity for the ability to learn I2I mapping with unpaired data. These methods employ GANs to adversarially align domain distribution, and meanwhile, preserve original image content via constraints like perceptual similarity \cite{gao2022learning}, cycle-consistency \cite{zhu2017unpaired}, geometry-consistency \cite{fu2019geometry}, and contrastive learning \cite{park2020contrastive}, etc. Later on, numerous methods have been proposed to improve UI2I in visual quality \cite{wang2022pretraining}, multimodality \cite{lee2020drit++}, multi-domain flexibility \cite{choi2020stargan}, efficiency \cite{zhang2022wavelet}, and few-shot learning capability \cite{pizzati2022manifest}. However, all these methods are only able to translate images between limited domains, unable to tackle open-domain I2I translation.

Another type of GAN-based I2I method is GAN inversion. Observing that well-trained StyleGAN \cite{karras2019style} encodes disentangled semantic information in latent space, these methods realize image semantic translation by inverting images into StyleGAN latent space and manipulating latent vectors. The inversion could be learned by online optimization \cite{wu2021stylespace}, training an encoder network \cite{richardson2021encoding}, or hybrid method \cite{roich2022pivotal}. Nonetheless, these methods rely on StyleGAN model pre-trained on a certain domain (e.g., human face domain) and thus are still domain-restricted. Therefore, they are less applicable to image translation with relatively large semantic gap.

Subsequently, research interest has been focused on leveraging CLIP \cite{radford2021learning} to guide I2I with text. VQCLIP \cite{crowson2022vqgan} optimizes VQGAN \cite{esser2021taming} latent vector under CLIP loss to translate an image as per a text. DiffusionCLIP \cite{kim2022diffusionclip} realizes text-guided image manipulation by fine-tuning pre-trained diffusion model to maximize image-text CLIP similarity. DiffuseIT  \cite{kwon2022diffusion} uses the derivative of the CLIP loss to align the diffusion model's reverse sampling process towards the target semantics described by the text. Under the same CLIP supervision, Text2LIVE \cite{bar2022text2live} trains a generator network that takes in the source image and target text and outputs an editing layer, which is composited over the source image for image editing. These CLIP-based methods allow using free-form text to instruct image translation, extending I2I from limited domains to open-domain capability. However, these methods are relatively slow and less efficient due to a separate CLIP-based optimization process for each time of image translation. Besides, the composition of multiple objective functions of these methods makes them tuning-unfriendly.

As large-scale text-to-image (T2I) diffusion models revolutionize the field of generative AI, methods have been proposed to harness their immense generative power and adapt them to the realm of text-guided I2I translation, where the key ingredient is to establish spatial structure consistency between the source image and the translated image. For example, SDEdit \cite{meng2021sdedit} preserves overall image structure by noising the source image to an intermediate diffusion step and then denoising it conditioned on the text prompt. Instructpix2pix \cite{brooks2023instructpix2pix} directly trains a text-driven I2I mapping in a supervised manner based on an elaborately synthesized large-scale paired dataset, where I2I structure consistency is naturally learned from training data. Prompt-to-Prompt \cite{hertz2022prompt} replaces the text-to-image cross-attention maps in the T2I denoising sampling process with the corresponding attention maps of the source image to ensure I2I structure similarity. Similarly, Plug-and-Play \cite{tumanyan2023plug} injects feature maps and self-attention maps of the source image into the T2I denoising sampling process to make the generated image comply with the source image in spatial structure. 

Despite stunning success of these methods, we observe that I2I has diverse application scenarios emphasizing different correlations (e.g., style, structure, layout, contour, etc.) between the source image and the translated image, and it is difficult for a single existing model to suit all I2I scenarios well. To this end, this paper proposes a unified framework suitable for diverse I2I applications from a novel frequency-domain perspective. Based on the intuition that I2I correlations like style, structure, layout, and contour are much more decomposable in the frequency domain than in the spatial domain, we propose to associate different I2I correlations to different spectral bands of image features in the Discrete Cosine Transform (DCT) spectral space. Specifically, our method filters image features in the DCT domain and uses the filtered image features bearing different DCT spectral bands as corresponding control signals to control different I2I correlations, and thus flexibly suits versatile I2I applications. Below we summarize various I2I application scenarios, the corresponding I2I correlations, and their relations to DCT spectral bands:

\begin{itemize}
    \item \textbf{Image style translation} aims to alter the global style of an image while maintaining the original fine structures (e.g., contours, edges). The fine structures to be preserved correspond to regions of high frequency in the DCT spectrum, which we term \emph{high-frequency spectral band}.
    \item \textbf{Style-guided content creation} aims to recreate arbitrary image content while preserving the original style information. The image style mainly manifests in color and luminance features that correspond to regions of especially low frequency in the DCT spectrum, which we term \emph{mini-frequency spectral band}.
    \item \textbf{Image semantic manipulation} aims to manipulate image semantic features without altering its global style and spatial structure. The preserved ``style $+$ structure'' corresponds to the low-frequency region in the DCT spectrum which has a wider bandwidth than the mini-frequency spectral band, we term it \emph{low-frequency spectral band}. 
    \item \textbf{Image scene translation} aims to translate images to a larger extent: convert the scene of an image to another one where only I2I layout similarity is pursued. To disentangle low-frequency image style and high-frequency image contours, we represent pure image layout information with the middle-frequency region in the DCT spectrum, i.e., \emph{mid-frequency spectral band}. 
\end{itemize}

Accordingly, we propose frequency-controlled diffusion model (FCDiffusion) for text-guided I2I. FCDiffusion basically adopts ControlNet \cite{zhang2023adding} paradigm which trains a network to control the T2I denoising sampling process of the pre-trained Latent Diffusion Model \cite{rombach2022high}, where the control signal here is the frequency-domain filtered latent features of the source image, which bear a certain DCT spectral band. Conditioned on the control signal, the model is trained to reconstruct the removed spectral components of the source image latent features via the textual information of the paired text prompt. At inference time, the source image can be translated by filling the missing DCT spectral region with the textual information of an arbitrary editing text prompt. As Fig. \ref{fig:cover} displays, our model can flexibly handle diverse I2I application scenarios under different modes of frequency control. The advantages of FCDiffusion are threefold: (\romannumeral1) it suits versatile I2I tasks simply by applying different DCT filters to construct the corresponding control signals; (\romannumeral2) it integrates multiple and scalable frequency control branches, enabling flexible switching among diverse I2I applications within a single model; (\romannumeral3) it is concise in learning objective, low-demanding in computational resources, efficient in inference speed, and competitive in I2I visual quality.

\begin{figure*}
    \centering
    \includegraphics[width=\textwidth]{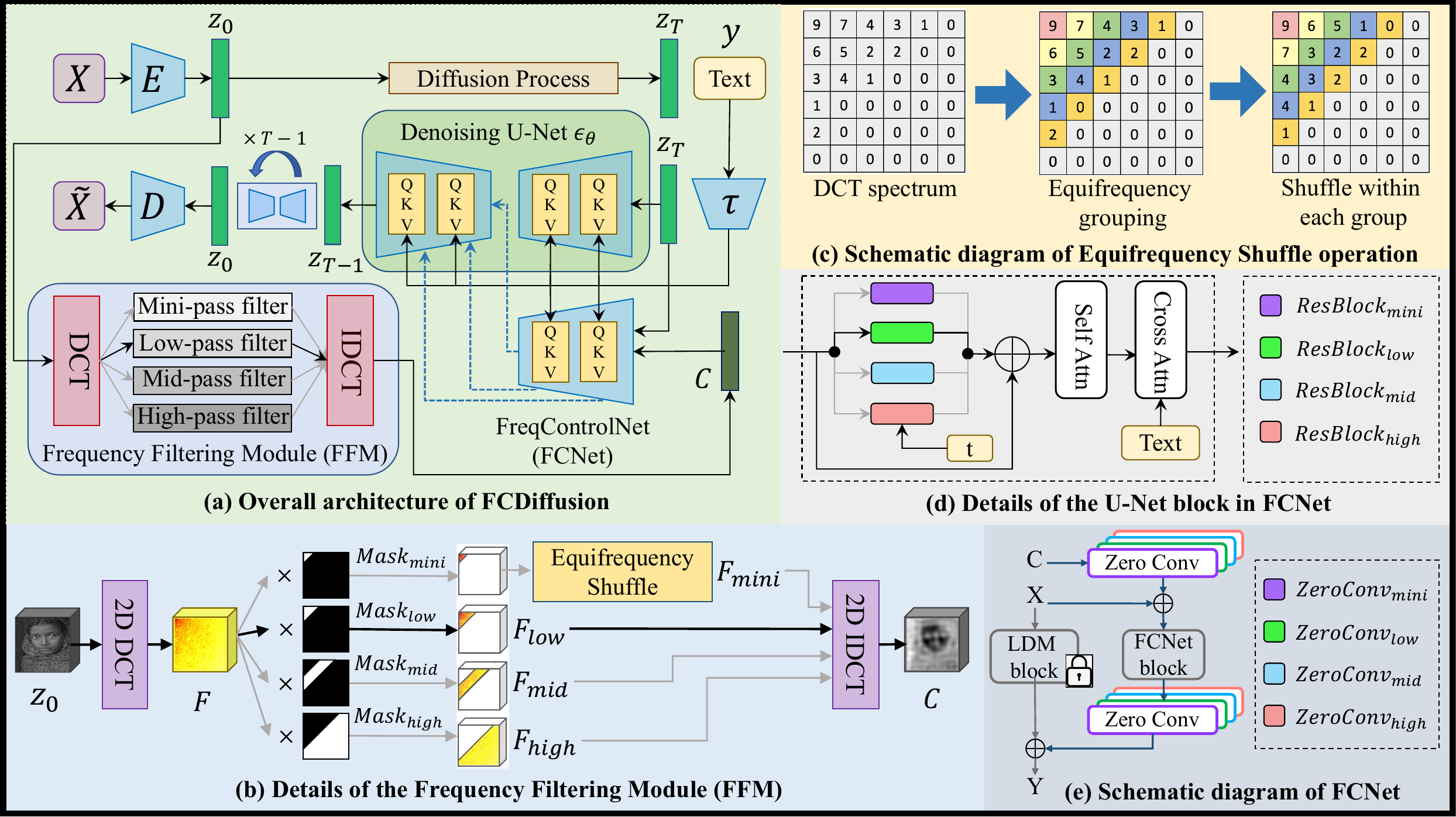}
    \caption{Overall architecture of FCDiffusion, as well as details of important model components.}
    \label{fig:method}
\end{figure*}

\section{Related Work}
\subsection{Diffusion Models}
With the advent of DDPM \cite{ho2020denoising}, diffusion models have received tremendous attention and have soon dominated the field of image generation \cite{dhariwal2021diffusion}. Then, efforts have been made to explore their potential in various vision problems such as super-resolution \cite{saharia2022image}, I2I \cite{saharia2022palette}, image inpainting \cite{lugmayr2022repaint}, etc. Boosted by vision-language multimodal technologies, large-scale text-to-image diffusion models (e.g., Imagen \cite{saharia2022photorealistic}, DALLE2 \cite{ramesh2022hierarchical}, GLIDE \cite{nichol2022glide}) impressively promote the prosperity of AIGC industry. Soon afterward, Latent Diffusion Model (LDM) \cite{rombach2022high} enables synthesizing high-resolution images with remarkably lower computational overhead by transferring diffusion process onto the low-dimensional feature space. ControlNet \cite{zhang2023adding} further facilitates controllable image generation by training a network to control LDM conditioned on certain image priors like Canny edge maps. Besides, diffusion models are also increasingly applied in vision fields like point cloud generation \cite{luo2021diffusion}, video synthesis \cite{mei2023vidm}, 3D reconstruction \cite{anciukevivcius2023renderdiffusion}, etc. 

\begin{figure*}
    \centering
    \includegraphics[width=\textwidth]{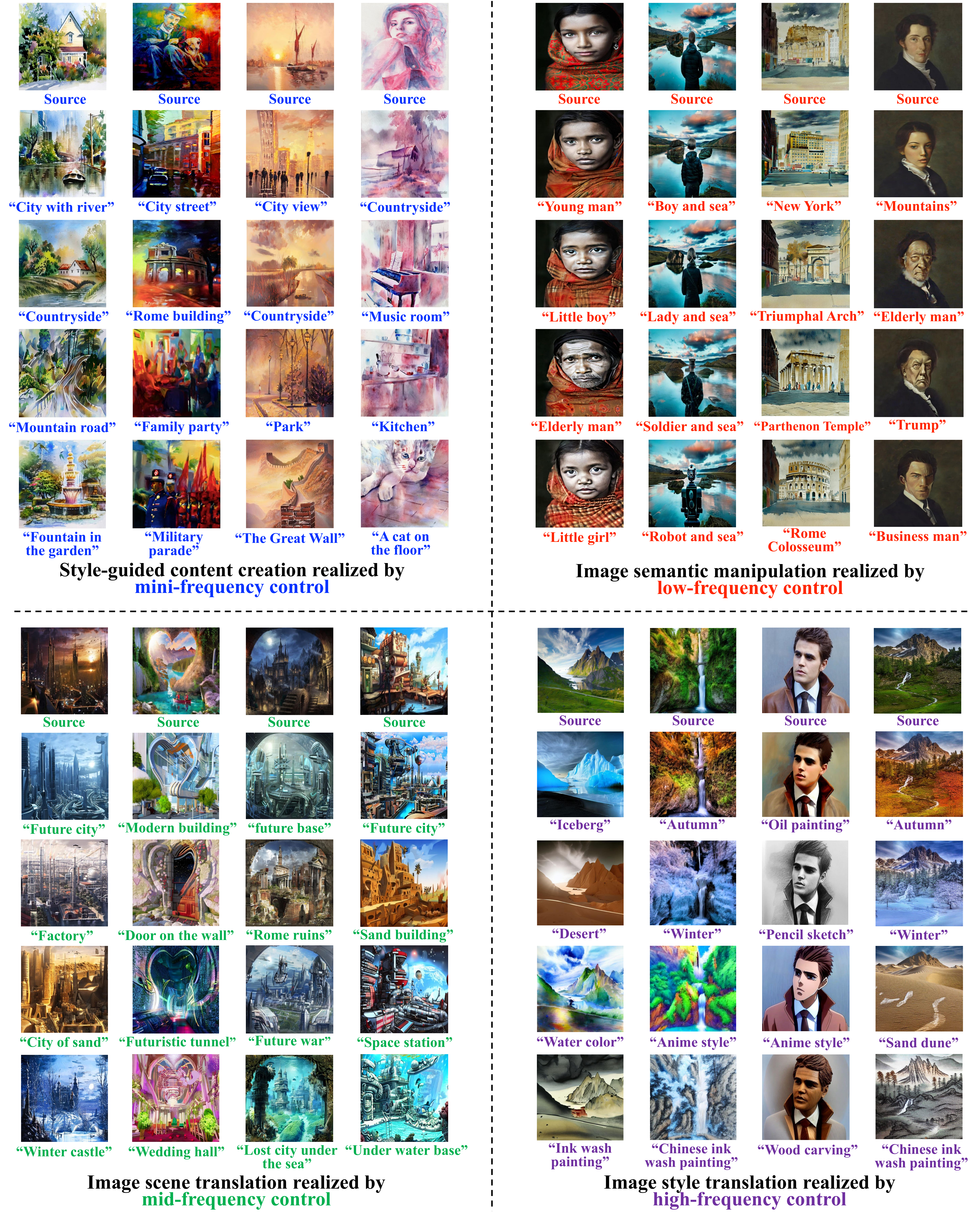}
    \caption{Example text-guided I2I results of our method. Our method suits diverse I2I application scenarios emphasizing different I2I correlations simply by switching to different modes of frequency control. The mini-frequency, low-frequency, mid-frequency, and high-frequency control respectively correlates the source image and the generated image in style, style and structure, layout, and contours, realizing style-guided content creation, image semantic manipulation, image scene translation, and image style translation, respectively. \textbf{Better viewed with zoom-in}.}
    \label{fig:samples}
\end{figure*}

\subsection{Deep Learning in Frequency Perspective}
Though neural networks are mostly exploited in spatial or temporal domains, some research work suggests that deep models can be improved from frequency-domain perspective. Ghosh et al. \cite{ghosh2016deep} accelerate CNN convergence by applying DCT operation on CNN feature maps. DCT-Conv \cite{chkeinski2020dct} combines convolution with DCT inverse transformation (IDCT) to form a novel network layer that shows advantage in network pruning. Xie et al. \cite{xie2021learning} propose a frequency-aware dynamic network which introduces DCT to image super-resolution model to lower computation overhead. Cai et al. \cite{cai2021frequency} propose to regulate image translation tasks with Fourier frequency spectrum consistency constraint, realizing better content preservation ability. This paper applies DCT filtering to the field of text-guided I2I, we realize versatile I2I applications with a single model by extracting image features bearing different DCT spectral bands to control different I2I correlations.

\section{Method}
In this section, we first introduce the overall model architecture, then elaborate on important modules, and finally describe the learning objective and training details.
\subsection{Overall Architecture}
As illustrated in Fig. \ref{fig:method}(a), FCDiffusion basically comprises three components: (\romannumeral1) pre-trained LDM, (\romannumeral2) Frequency Filtering Module (FFM), (\romannumeral3) FreqControlNet (FCNet).

The pre-trained LDM uses a strong auto-encoder to compress a source image $X \in \mathbb{R}^{H\times W\times 3}$ to a compact latent representation $z_{0} \in \mathbb{R}^{h\times w \times c}$, i.e., $z_{0}=E(X)$, $D(z_{0})=D(E(X))\approx X$, where $\frac{H}{h}=\frac{W}{w}=8$, $c=4$. A DDPM is trained at feature space to recover $z_{0}$ from Gaussian distribution conditioned on the paired text prompt $y$: 
\begin{equation}
    L_{LDM}=\mathbb{E}_{z_{0}, y, \epsilon\sim\mathcal{N}(0, 1), t}[\left\|\epsilon-\epsilon_{\theta}(z_{t}, t, \tau(y))\right\|_{2}^{2}],
\end{equation}
where $t$ denotes a sampled time step, $z_{t}$ is the noised feature at time step $t$, $\epsilon_{\theta}$ is the denoising U-Net which takes $z_{t}$, time step $t$, and text embedding $\tau(y)$ as input and outputs the estimation of the Gaussian noise sampled in the forward diffusion process, $\tau$ is the OpenCLIP transformer text encoder. We omit the technical details of DDPM and LDM here since they are not relevant to our key contributions. 

To adapt LDM from text-to-image generation to text-guided I2I, a Frequency Filtering Module (FFM) is constructed to filter the encoded image features $z_{0}$ of the source image $X$ in the frequency domain (DCT spectrum), the filtered image features $C=FFM(z_{0})$ function as a control signal which controls the reverse denoising sampling process of the LDM through a FreqControlNet (FCNet). The FCNet takes in the control signal $C$ that contains only partial frequency spectrum components (a certain DCT spectral band) of $z_{0}$ and is optimized towards guiding the LDM to reconstruct  $z_{0}$ with the partial spectral information from $C$ and the textual information from $y$. From the frequency-domain perspective, the training of FCNet can be regarded as a process of recovering the removed frequency spectrum components of $z_{0}$ via the paired text prompt $y$.

\subsection{Frequency Filtering Module}
As detailed in Fig. \ref{fig:method}(b), in FFM, channel-wise 2D DCT is firstly applied to convert the source-domain latent features $z_{0}$ into the frequency-domain counterpart $F$: 
\begin{equation}
\begin{aligned}
    &F^{(n)}_{u,v}=2/(\sqrt{hw})m(u)m(v)\sum\nolimits_{i=0}^{h-1}\sum\nolimits_{j=0}^{w-1}[(z_{0}^{(n)})_{i,j} \\
    &\cos((2i+1)u\pi/(2h))\cos((2j+1)v\pi/(2w))],
\end{aligned}
\end{equation}

where $m(0)=\frac{1}{\sqrt{2}}$, $m(\gamma)=1$ for all $\gamma>0$, $F^{(n)}$ and $z_{0}^{(n)}$ are the $n^{th}$ channel of $F$ and $z_{0}$ respectively, $n=1, 2, ..., c$. The index $i$, $j$ denote the 2D coordinate in the spatial domain, while $u$, $v$ refer to the 2D coordinate in the DCT frequency domain. In 2D DCT spectrum, elements with smaller coordinates (nearer to the top-left origin) encode lower-frequency information, while larger-coordinate elements (nearer to the bottom-right corner) correspond to higher-frequency components. As summarized in the Introduction part, different spectral bands in the DCT domain encode different image visual attributes and thus can be used as conditional guidance to control different I2I correlations. Therefore, we manually design four DCT filters (i.e., DCT masks) for mini-pass, low-pass, mid-pass, and high-pass frequency-domain filtering respectively, as described in detail below:
$$\left\{ 
        \begin{array}{lr}
             Mask_{mini}(u,v)=1\ \ if\ \ u+v\leq10\ \ else\ \ 0, &  \\
             Mask_{low}(u,v)=1\ \ if\ \ u+v\leq20\ \ else\ \ 0, &  \\
             Mask_{mid}(u,v)=1\ \ if\ \ 20<u+v\leq40\ \ else\ \ 0, &  \\
             Mask_{high}(u, v)=1\ \ if\ \ u+v\geq50\ \ else\ \ 0.
        \end{array}
\right.
$$
These DCT filters extract DCT features containing only the mini-frequency, low-frequency, mid-frequency, and high-frequency spectral band respectively, which are implemented via direct multiplication with $F$:
\begin{equation}
    F_{*}=F\times Mask_{*},
\end{equation}
where $*$ $\in$ \{mini, low, mid, high\}. Finally, 2D IDCT is used to convert the filtered DCT features $F_{*}$ back to the spatial domain as the final control signal $C$:
\begin{equation}
\begin{aligned}
    &C^{(n)}_{i,j}=2/(\sqrt{hw})\sum\nolimits_{u=0}^{h-1}\sum\nolimits_{v=0}^{w-1}[m(u)m(v)(F_{*}^{(n)})_{u,v} \\
    &\cos((2i+1)u\pi/(2h))\cos((2j+1)v\pi/(2w))],
\end{aligned}
\end{equation}
where $F_{*}^{(n)}$ and $C^{(n)}$ are the $n^{th}$ channel of $F_{*}$ and $C$ respectively, $n=1, 2, ..., c$. The control signal $C$ extracted with the mini-pass, low-pass, mid-pass, and high-pass DCT filtering respectively controls the style, style and structure, layout, and contour consistency between the generated image and the source image, allowing diverse I2I application scenarios emphasizing different I2I correlations.

\begin{figure*}[t]
    \centering
    \includegraphics[width=\textwidth]{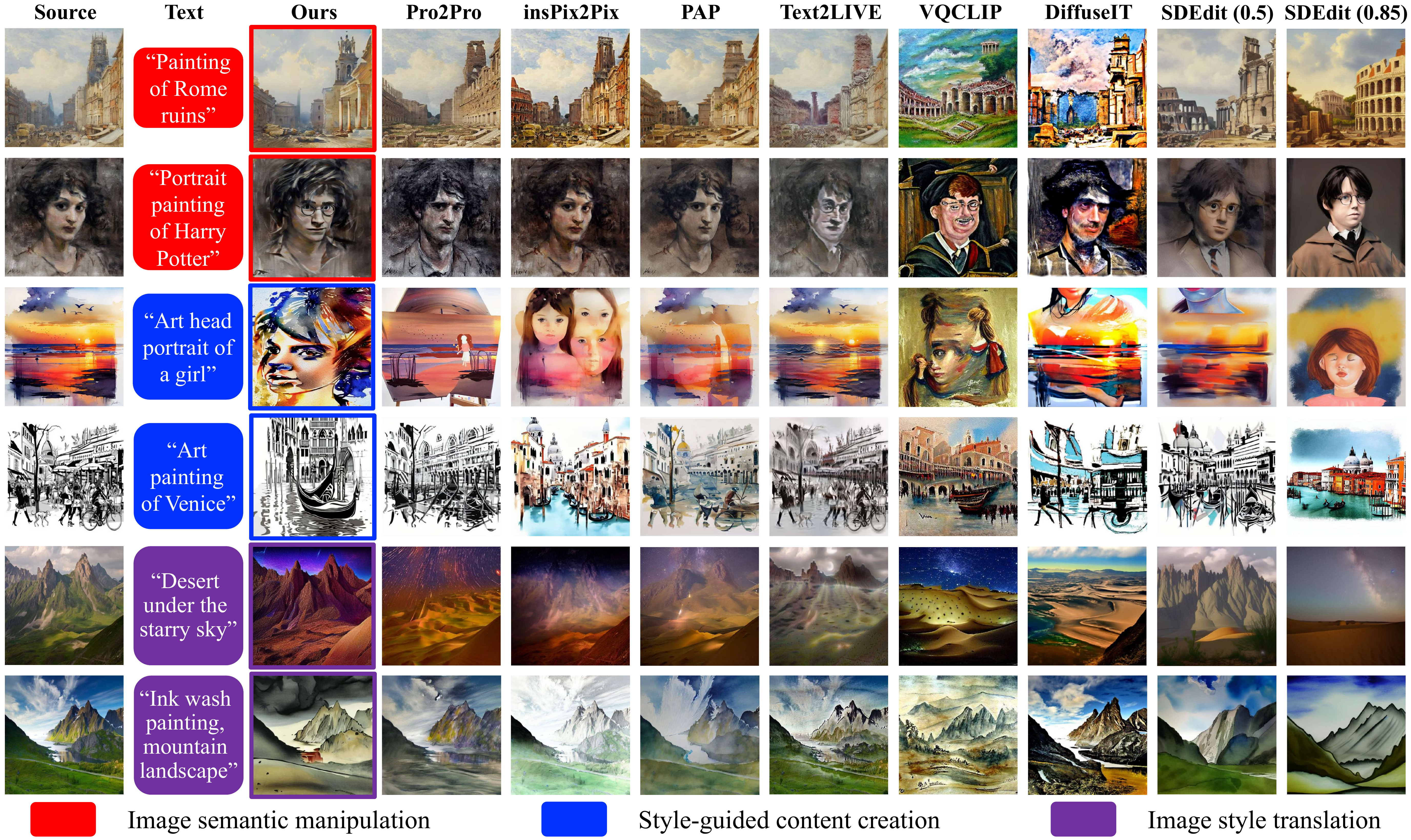}
    \caption{Visual comparisons of our method with related text-guided image translation methods on different I2I tasks including image semantic manipulation (top two rows), style-guided content creation (middle two rows), and image style translation (bottom two rows). Results of our method for these three tasks are obtained by switching to the low-frequency, mini-frequency, and high-frequency control branch respectively. \textbf{Better viewed with zoom-in}.}
    \label{fig:method_compare}
\end{figure*}

Empirically, we observe that the control signal after the mini-pass filtering may still contain some image structural information and thus will still weakly constrain the global structure of the generated image. To eliminate structural constraint and achieve pure style control, we propose and append an Equifrequency Shuffle operation to the end of the mini-pass filtering branch at inference time. As Fig. \ref{fig:method}(c) illustrates, the operation firstly groups DCT components by the frequency level which is quantified as the sum of 2D coordinates in the DCT spectrum, then shuffles DCT elements within each group. The Equifrequency Shuffle randomly perturbs DCT spectrum without altering its overall energy distribution, which is useful to eliminate the spatial structure influence of the source image to the generated image, all while maintaining the I2I style correlation.

\subsection{FCNet Architectural Details}
At each time step $t$, the FCNet takes the current denoising result $z_{t}$, together with the control signal $C$ and the text embedding $\tau(y)$ as input, and outputs multi-scale feature maps to guide the pre-trained LDM towards reconstructing $z_{0}$. Borrowing from ControlNet \cite{zhang2023adding}, FCNet is a trainable copy of the LDM U-Net encoder. As Fig. \ref{fig:method}(d) displays, each U-Net block in the FCNet consists of a ResBlock that incorporates the time embedding, a self-attention block, and a cross-attention block that incorporates the text embedding. Each ResBlock in the FCNet has four parallel replicates corresponding to the four DCT filtering branches in the FFM. As Fig.\ref{fig:method}(e) illustrates, the Zero Convolutions proposed in ControlNet \cite{zhang2023adding} are also utilized here for smooth feature injection into the pre-trained LDM. Similarly, each Zero Convolution also has four parallel replicates corresponding to the four DCT filtering branches.

\subsection{Learning Objective and Training Details}
Our framework is fully differentiable and end-to-end trainable. The objective is to reconstruct the source image latent features $z_{0}$ with the control signal $C=FFM(z_{0})$ which bears only partial DCT spectral band of $z_{0}$, as well as the paired text prompt $y$. This is equivalent to minimizing the following conditional noise regression loss:
\begin{equation}
    L=\mathbb{E}_{z_{0}, y, \epsilon\sim\mathcal{N}(0, 1), t}[||\epsilon-\epsilon_{\theta}(z_{t}, t, \tau(y), c_{t})||_{2}^{2}],
\end{equation}
\begin{equation}
    c_{t}=FCNet(FFM(z_{0}), t, \tau(y)).
\end{equation}
The FCNet, as the only trainable module in the entire framework, is initialized from the pre-trained LDM except for the additional Zero Convolutions. In the training phase, we freeze all self-attention and cross-attention layers in the FCNet and only fine-tune ResBlocks and Zero Convolutions. We observe that freezing and sharing attention layers reduce trainable parameters without degrading generation performance. Our model contains four frequency control branches consisting of \textbf{[}Mask$_{*}$, ResGroup$_{*}$, ZeroGroup$_{*}$\textbf{]}, where $*$ $\in$ \{mini, low, mid, high\}. ResGroup$_{*}$ represents the parameter group which comprises all ResBlock$_{*}$ replicates in the FCNet that correspond to the DCT filter Mask$_{*}$, and the same for ZeroGroup$_{*}$. These four frequency control branches are separately fine-tuned and can be flexibly switched at inference time for different I2I application scenarios. It is worth mentioning that the control branches in our model can be scalable and pluggable, that is, more I2I control effects can be realized simply by designing the corresponding DCT filters and allocating additional ResGroup and ZeroGroup for fine-tuning. 

\begin{figure*}[t]
    \centering
    \includegraphics[width=\textwidth]{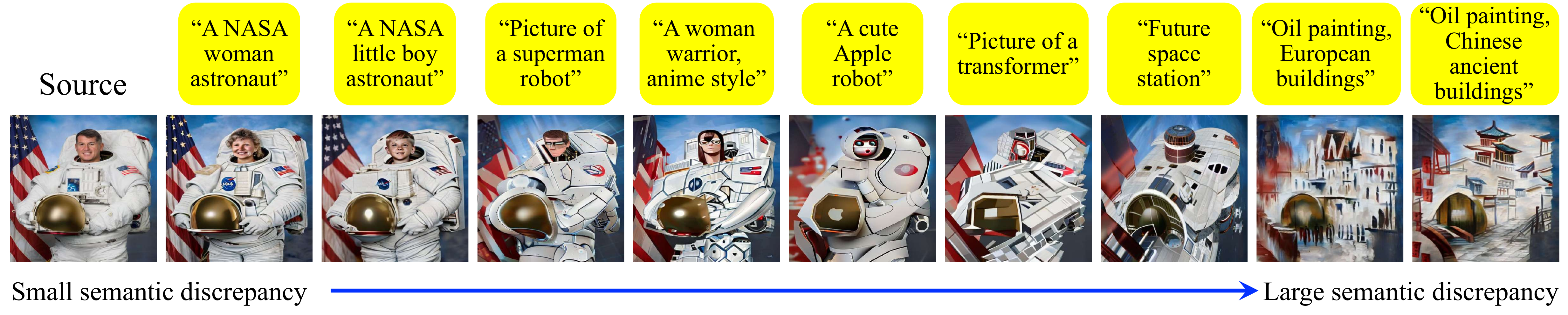}
    \caption{With low-frequency control, our method is able to manipulate image semantics under different degrees of semantic discrepancy. As the semantic gap between the source image and the target text increases, the translated image can still conform to the text with the original image style and structure preserved. Better viewed with zoom-in.}
    \label{fig:semantic_adaptation}
\end{figure*}

\section{Experiments}
\subsection{Experiment Setup}
We use Stable Diffusion v2-1-base as the pre-trained LDM in our method, and use LAION-Aesthetics 6.5$+$ which contains 625K image-text pairs as our dataset, in which we randomly partition into a training set and a test set at the ratio of 9:1. We train at 512$\times$512 image resolution, i.e., $H=W=512, h=w=64$. We set the initial learning rate as 1e-5. Each frequency control branch in our model is separately fine-tuned for 100K iterations with batch size 4 on a single RTX 3090 Ti GPU. At inference time, FCDiffusion can flexibly adapt to versatile I2I application scenarios simply by switching among different frequency control branches. All the results in this paper are generated using the DDIM \cite{song2020denoising} sampler with 50 sampling steps.

\begin{figure}
    \centering
    \includegraphics[width=3.3in]{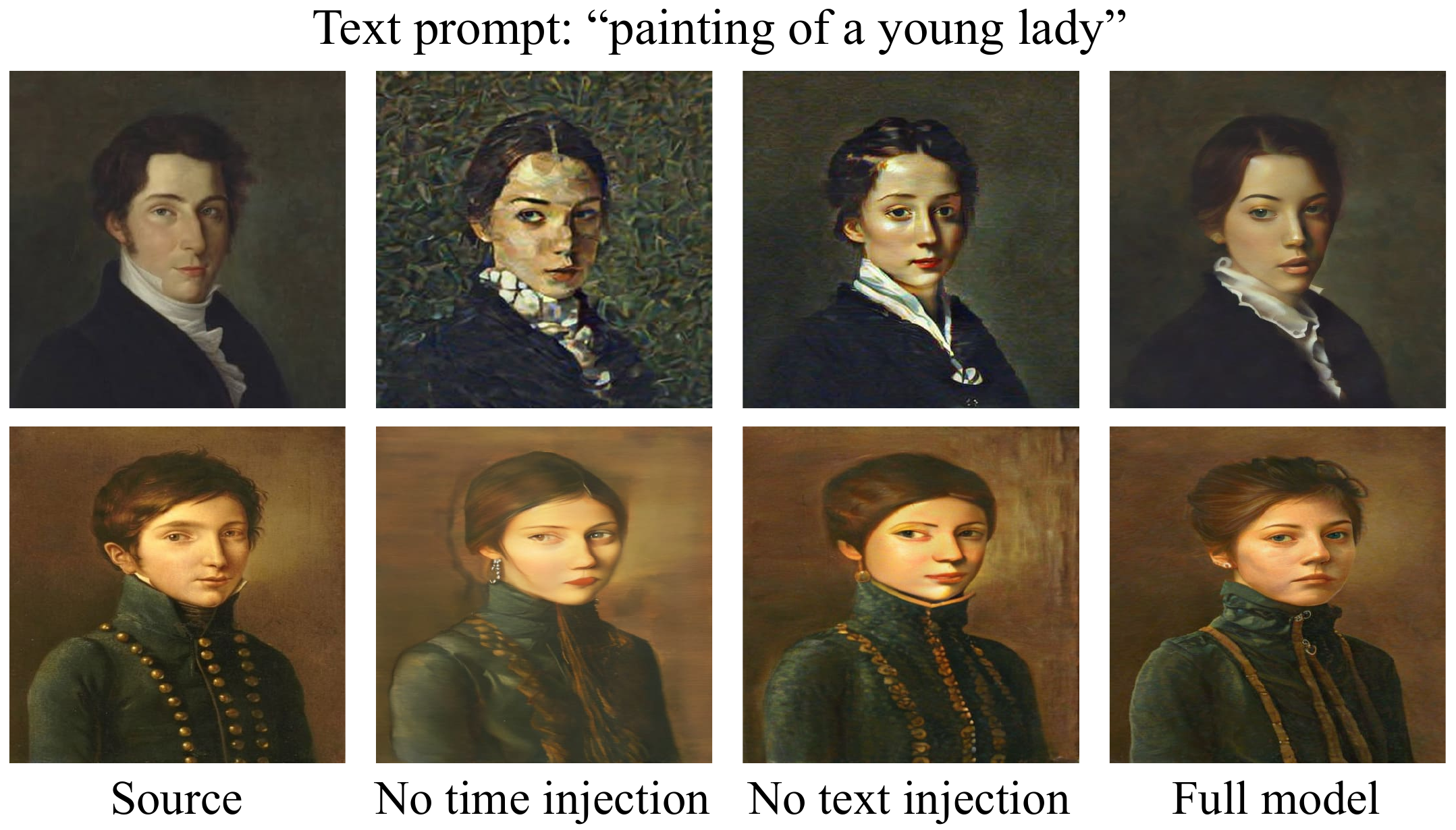}
    \caption{Qualitative ablation study on the architectural design of the FCNet. \textbf{Better viewed with zoom-in}.}
    \label{fig:ablation1}
\end{figure}

\subsection{Qualitative Analyses}
As Fig. \ref{fig:samples} displays, our method produces high-quality results in diverse I2I scenarios by switching among different frequency control branches. With mini-frequency control, the translated image only preserves the original style information without structural constraint of the source image, and thus realizes the application of style-guided content creation, i.e., recreate any image content as per the text prompt without altering image style. Under the low-frequency control, both image style and spatial structure of the source image are maintained to the generated image, which suits smaller-scale editing of the source image, i.e., image semantic manipulation. Under the high-frequency control, the translated image conforms to the source image in object contours with little constraint on style appearance, which allows to manipulate image style according to the text prompt, i.e., image style translation. Besides, we also realize application of image scene translation where the only I2I correlation is image layout, for which we resort to the mid-frequency control to bypass constraints from the source image in low-frequency style and high-frequency contours.

\begin{figure}
    \centering
    \includegraphics[width=3.3in]{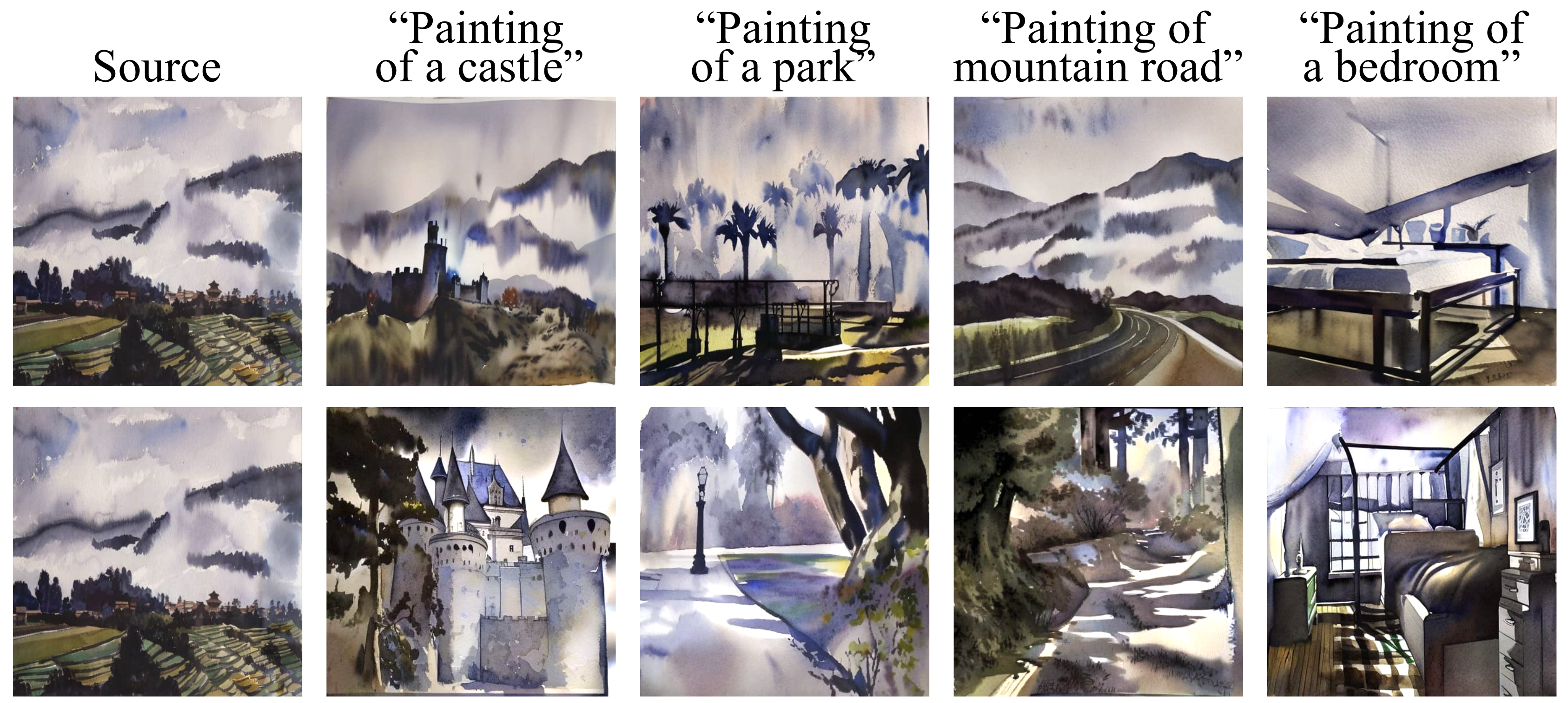}
    \caption{Qualitative ablation study on the style-guided content creation task \textbf{without} (top row) and \textbf{with} (bottom row) the \textbf{Equifrequency Shuffle} operation. \textbf{Better viewed with zoom-in}.}
    \label{fig:ablation2}
\end{figure}

In Fig. \ref{fig:method_compare}, we qualitatively compare our method with related advanced methods on some challenging I2I examples. We abbreviate methods of Prompt-to-Prompt, InstructPix2Pix, Plug-and-Play as Pro2Pro, insPix2Pix, PAP respectively for brevity. For image semantic manipulation, Pro2Pro, insPix2Pix, and PAP are less able to generate semantically faithful results while achieving high consistency in style appearance and spatial structure. Results of Text2LIVE contain severe artifacts. VQCLIP and DiffuseIT fail to preserve original style distribution of the source image. In the track of style-guided content creation, Pro2Pro struggles to generate image content that is faithful to the text prompt. Results of insPix2Pix better comply with the text prompt in content semantics but are relatively weak in style preservation. Other methods either fail to generate faithful content or fail to maintain original style appearance. For image style translation, results of VQCLIP manifest accurate style distribution instructed by the text but lose original contour information of the source image. Text2LIVE, on the contrary, preserves image contours well but fails to precisely translate image style as per the text prompt. Other methods either fail to maintain contour consistency or cannot translate image style faithfully. 

\begin{figure*}[t]
    \centering
    \includegraphics[width=0.85\textwidth]{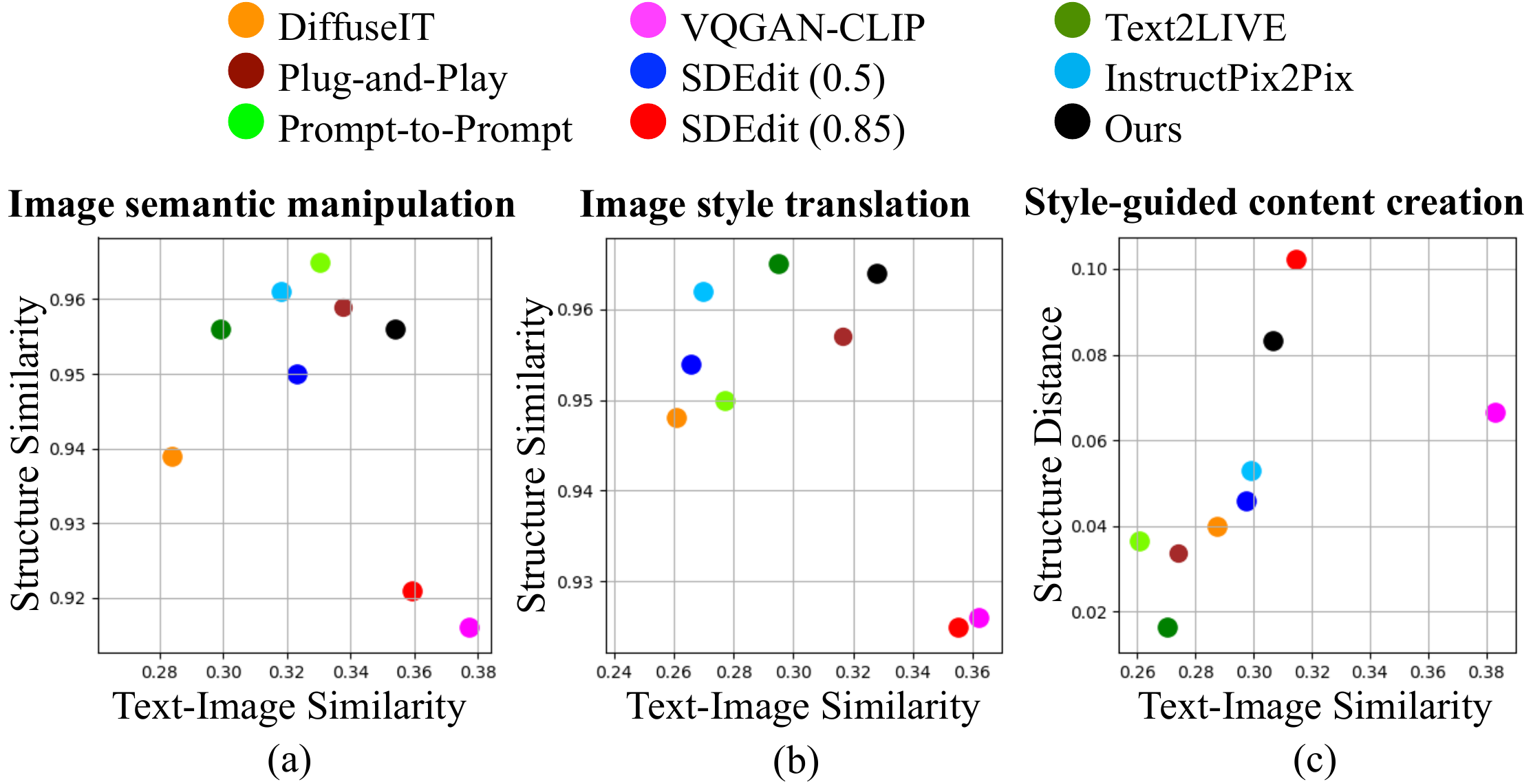}
    \caption{Quantitative comparison of our method to related approaches on different text-guided I2I tasks, including image semantic manipulation, image style translation, and style-guided content creation, where the former two tasks prefer high text fidelity and high structure similarity, and the last task favor high text fidelity and large structure distance (low structure similarity). Our method achieves competitive performance (near the top-right corner) in all the three tasks.}
    \label{fig:quantitative_results}
\end{figure*}

\begin{figure}[t]
    \centering
    \includegraphics[width=3.3in]{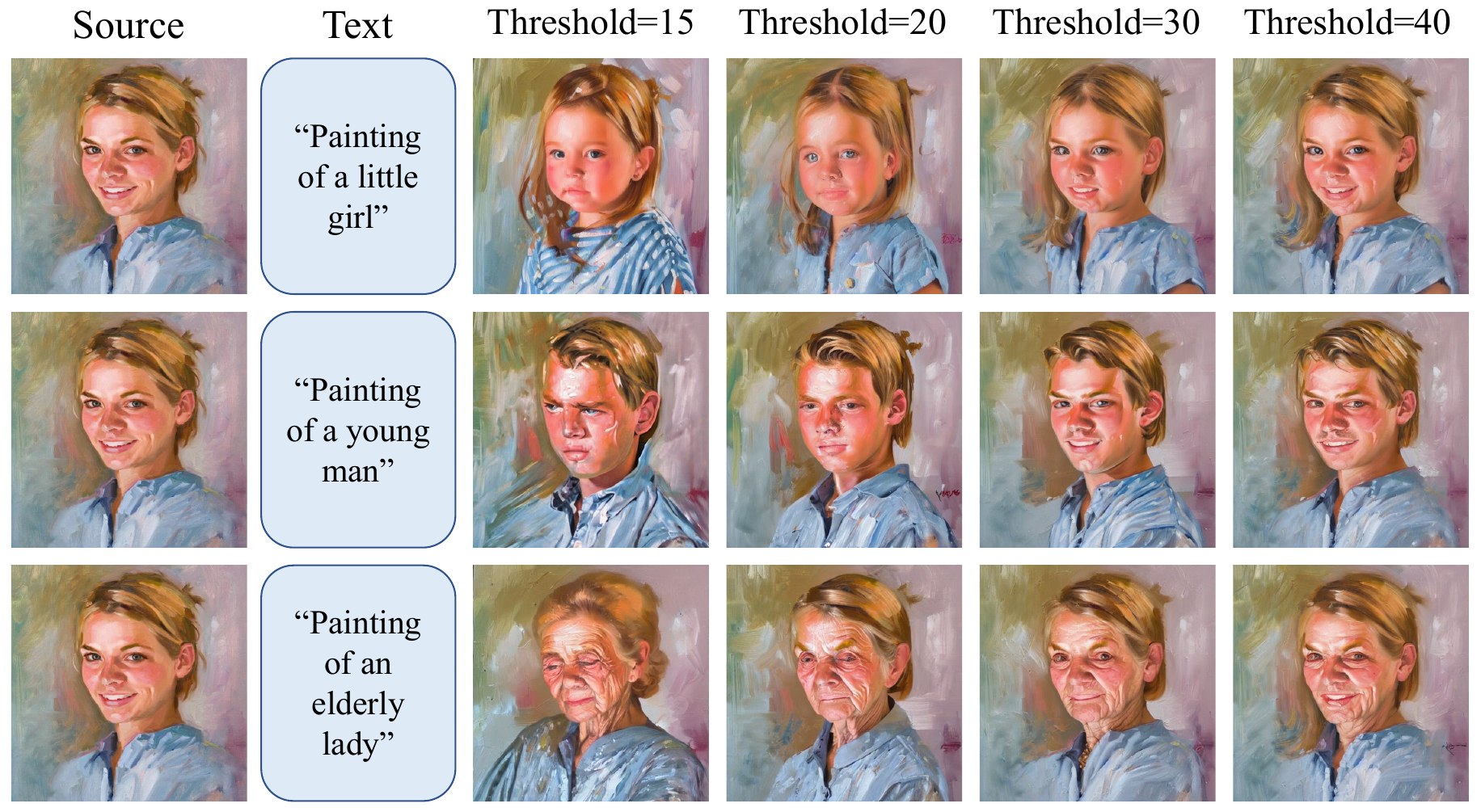}
    \caption{Example image semantic manipulation results of our method under the low-frequency control with varying values of the low-pass filtering threshold.}
    \label{fig:band_range_ablation}
\end{figure}

Besides, we also compare to SDEdit with different noising strength values (shown in the parentheses). SDEdit(0.5) can basically preserve spatial structure of the source image, but is less effective in maintaining style appearance for image semantic manipulation, creating structure-invariant content for style-guided content creation, and altering image style sufficiently for image style translation. For large noising strength, results of SDEdit(0.85) suffer from weak connection to the source image, losing the corresponding I2I correlation in all I2I tasks. By contrast, our method effectively achieves the desired performance with high quality for all I2I applications.

As demonstrated in Fig. \ref{fig:semantic_adaptation}, FCDiffusion can handle image semantic manipulation with not only narrow semantic gaps but also large semantic discrepancies between the source image and the target text prompt, based on the low-frequency control of our method. The translated images can still comply with the text in semantics and preserve the original ``style$+$structure'' of the source image simultaneously even if the target text and the source image are semantically uncorrelated.

\subsection{Ablation Studies}
We ablate our model design from the following four aspects:
(\romannumeral1) the necessity of injecting time embedding into FCNet; (\romannumeral2) the necessity of injecting text embedding into FCNet; (\romannumeral3) the effectiveness of our Equifrequency Shuffle operation in decorrelating image spatial structure between the source image and the translated image; (\romannumeral4) the influence of the frequency band range to text-guided I2I. We conduct the first study by removing all layers that incorporates time embedding from the FCNet. For the second study, we do not modify the cross-attention layers but instead feed the null text to the FCNet during training. As qualitatively displayed in Fig. \ref{fig:ablation1}, removing time embedding injection in the FCNet yields unstable image translation results with noticeable noises and artifacts, which means that it is important for FCNet to learn to provide time-dependent guidance features to the LDM. Besides, removing text embedding injection in the FCNet also leads to results with lower image quality, which indicates that textual information is useful for the FCNet in providing finer control to the LDM. Example results of the third study are shown in Fig. \ref{fig:ablation2}, from which we see that the style-guided content creation results obtained without Equifrequency Shuffle still resemble the source image in global structure to a certain degree, whereas results obtained with Equifrequency Shuffle are much more disentangled from the source image in spatial structure, which demonstrates the effectiveness of this operation in decorrelating spatial structure in the style-guided content creation task. For the last study, we qualitatively compare the image semantic manipulation results of our method achieved under the low-frequency control with varying low-pass filtering thresholds (i.e., varying bandwidth of the extracted low-frequency spectral band). Results displayed in Fig. \ref{fig:band_range_ablation} show that the higher the threshold is, the closer the translated image is to the source image. Intuitively, raising the low-pass filtering threshold results in the control signal $C$ that bears a wider DCT spectral band of the source-image latent features. Thus, the preserved source-image information is increased and the translated image more resembles the source image. Conversely, lowering the low-pass filtering threshold leads to less spectral information of the source image in the control signal $C$. In this case, less source-image information is maintained, and thus less resemblance of the translated image to the source image.

\begin{table}[t]
            \large
		\caption{Inference speed comparison among different text-guided I2I methods for translating a single 512 $\times$ 512 image. Results are evaluated on an NVIDIA RTX 3090 Ti GPU and averaged on 10 trials for each method.} 
		\label{tab:inference_speed}
		\centering
            \resizebox{\linewidth}{!}{
            \begin{tabular}{cccc} 
			\toprule 
			Method & Inversion & Online optimization & Inference time\\
			\midrule
			VQCLIP & \XSolidBrush & \CheckmarkBold & 237.7s\\
                DiffuseIT & \XSolidBrush & \CheckmarkBold & 125.4s\\
                Text2LIVE & \XSolidBrush & \CheckmarkBold & 474.6s\\
                insPix2Pix & \XSolidBrush & \XSolidBrush & 11.2s\\
                Prompt-to-Prompt & \CheckmarkBold & \CheckmarkBold & 69.7s\\
                Plug-and-Play & \CheckmarkBold & \XSolidBrush & 102.9s\\
                FCDiffusion (Ours)& \XSolidBrush & \XSolidBrush & \textbf{9.4s}\\
			\midrule
		\end{tabular}
            }
	\end{table}
 
\subsection{Quantitative Evaluations}
For quantitative evaluation, we conduct 200 text-guided image translations for each I2I application scenario and compute the average value of the text-image similarity score and the I2I structure distance. The text-image similarity score is used to measure the fidelity of the translated image to the target text prompt (i.e., text fidelity), for which we use CLIP cosine similarity \cite{radford2021learning} as the metric. The I2I structure distance aims to measure the spatial structure discrepancy between the source image and the translated image, for which we use DINO-ViT self-similarity distance \cite{tumanyan2022splicing} as the metric. Correspondingly, we define I2I structure similarity as $1$ minus structure distance. 

For image semantic manipulation and image style translation tasks, methods are expected to achieve both high text fidelity and high I2I structure similarity. Though these two metrics contradict each other, our method achieves the most top-right position in Fig. \ref{fig:quantitative_results}(a) and \ref{fig:quantitative_results}(b), indicating the best trade-off of our results between text fidelity and structure similarity. For the task of style-guided content creation, methods are encouraged to generate spatial-structure-disentangled image translation results that comply with the text prompt in content descriptions. Therefore, high text fidelity and low structure similarity (i.e., large structure distance) are preferred in this scenario. As Fig. \ref{fig:quantitative_results} (c) shows, our method achieves competitive performance in both two aspects. Though SDEdit(0.85) and VQGAN-CLIP achieve more top-right position than our method in this task, these two methods suffer from very low structure similarity in the former two applications which favor I2I structure consistency, as shown in Fig. \ref{fig:quantitative_results}(a) and Fig. \ref{fig:quantitative_results}(b).

We further compare our method with related text-guided I2I approaches in inference speed by calculating the average running time of translating a single 512$\times$512 image. Results of different methods evaluated on an NVIDIA RTX 3090 Ti GPU are reported in Tab. \ref{tab:inference_speed}. CLIP-based methods such as VQCLIP, DiffuseIT, and Text2LIVE are significantly slow due to the required online optimization process driven by the CLIP loss to manipulate an image with a text. Inversion-based methods like Prompt-to-Prompt and Plug-and-Play spend most of the running time in DDIM inversion (diffusion trajectory to invert the source image into Gaussian noise space) and DDIM reconstruction (denoising trajectory to reconstruct the source image from the inverted noise feature), which makes image translation still time-consuming. On the contrary, our method and insPix2Pix are free from any inversion or online optimization process, and thus exhibit significant advantage in terms of inference speed.

\section{Conclusion and Discussion}
In this paper, we propose a solution to text-guided I2I from a novel frequency-domain perspective. The proposed model, FCDiffusion, adapts the pre-trained LDM from text-to-image generation to the realm of text-guided I2I, which is realized by filtering the source image features in the DCT domain and using the filtered image features bearing different DCT spectral bands to control the reverse diffusion process of the Latent Diffusion Model. Our method can be interpreted as learning to fill the removed spectral components in the DCT spectrum of the source-image latent features with the textual information from the paired text prompt. By designing different modes of DCT filters as well as allocating and training the corresponding control branches, our method realizes diverse control effects of the source image to the text-guided generated image, allowing versatile text-driven I2I applications emphasizing different I2I correlations, including style-guided content creation realized by mini-frequency control, image semantic manipulation realized by low-frequency control, image scene translation realized by mid-frequency control, and image style translation realized by high-frequency control, where the source image and the translated image are correlated in image style, image style and structure, image layout, and image contour, respectively. Our method efficiently suits versatile I2I tasks simply by switching among different control branches at inference time, all while bearing advantages in inference speed and generation quality. Despite the advantage in versatility, our method needs to allocate and train a new frequency control branch when designing a new DCT filter, which is still less flexible and efficient. In future work, we will explore training-free and plug-and-play spectral band control mechanisms that enable plugging in arbitrary DCT filters at inference time to achieve continuous control effects.

\bibliography{aaai24.bib}

\begin{thebibliography}{46}
\providecommand{\natexlab}[1]{#1}

\bibitem[{Anciukevi{\v{c}}ius et~al.(2023)Anciukevi{\v{c}}ius, Xu, Fisher, Henderson, Bilen, Mitra, and Guerrero}]{anciukevivcius2023renderdiffusion}
Anciukevi{\v{c}}ius, T.; Xu, Z.; Fisher, M.; Henderson, P.; Bilen, H.; Mitra, N.~J.; and Guerrero, P. 2023.
\newblock Renderdiffusion: Image diffusion for 3d reconstruction, inpainting and generation.
\newblock In \emph{Proceedings of the IEEE Conference on Computer Vision and Pattern Recognition}, 12608--12618.

\bibitem[{Bar-Tal et~al.(2022)Bar-Tal, Ofri-Amar, Fridman, Kasten, and Dekel}]{bar2022text2live}
Bar-Tal, O.; Ofri-Amar, D.; Fridman, R.; Kasten, Y.; and Dekel, T. 2022.
\newblock Text2live: Text-driven layered image and video editing.
\newblock In \emph{Proceedings of the European Conference on Computer Vision}, 707--723.

\bibitem[{Brooks, Holynski, and Efros(2023)}]{brooks2023instructpix2pix}
Brooks, T.; Holynski, A.; and Efros, A.~A. 2023.
\newblock Instructpix2pix: Learning to follow image editing instructions.
\newblock In \emph{Proceedings of the IEEE Conference on Computer Vision and Pattern Recognition}, 18392--18402.

\bibitem[{Cai et~al.(2021)Cai, Zhang, Huang, Geng, Li, and Huang}]{cai2021frequency}
Cai, M.; Zhang, H.; Huang, H.; Geng, Q.; Li, Y.; and Huang, G. 2021.
\newblock Frequency domain image translation: More photo-realistic, better identity-preserving.
\newblock In \emph{Proceedings of the IEEE International Conference on Computer Vision}, 13930--13940.

\bibitem[{Ch{\k{e}}i{\'n}ski and Wawrzy{\'n}ski(2020)}]{chkeinski2020dct}
Ch{\k{e}}i{\'n}ski, K.; and Wawrzy{\'n}ski, P. 2020.
\newblock DCT-Conv: Coding filters in convolutional networks with Discrete Cosine Transform.
\newblock In \emph{International Joint Conference on Neural Networks}, 1--6.

\bibitem[{Choi et~al.(2020)Choi, Uh, Yoo, and Ha}]{choi2020stargan}
Choi, Y.; Uh, Y.; Yoo, J.; and Ha, J.-W. 2020.
\newblock Stargan v2: Diverse image synthesis for multiple domains.
\newblock In \emph{Proceedings of the IEEE Conference on Computer Vision and Pattern Recognition}, 8188--8197.

\bibitem[{Crowson et~al.(2022)Crowson, Biderman, Kornis, Stander, Hallahan, Castricato, and Raff}]{crowson2022vqgan}
Crowson, K.; Biderman, S.; Kornis, D.; Stander, D.; Hallahan, E.; Castricato, L.; and Raff, E. 2022.
\newblock Vqgan-clip: Open domain image generation and editing with natural language guidance.
\newblock In \emph{Proceedings of the European Conference on Computer Vision}, 88--105.

\bibitem[{Dhariwal and Nichol(2021)}]{dhariwal2021diffusion}
Dhariwal, P.; and Nichol, A. 2021.
\newblock Diffusion models beat gans on image synthesis.
\newblock \emph{Advances in Neural Information Processing Systems}, 34: 8780--8794.

\bibitem[{Esser, Rombach, and Ommer(2021)}]{esser2021taming}
Esser, P.; Rombach, R.; and Ommer, B. 2021.
\newblock Taming transformers for high-resolution image synthesis.
\newblock In \emph{Proceedings of the IEEE Conference on Computer Vision and Pattern Recognition}, 12873--12883.

\bibitem[{Fu et~al.(2019)Fu, Gong, Wang, Batmanghelich, Zhang, and Tao}]{fu2019geometry}
Fu, H.; Gong, M.; Wang, C.; Batmanghelich, K.; Zhang, K.; and Tao, D. 2019.
\newblock Geometry-consistent generative adversarial networks for one-sided unsupervised domain mapping.
\newblock In \emph{Proceedings of the IEEE Conference on Computer Vision and Pattern Recognition}, 2427--2436.

\bibitem[{Gao, Zhang, and Tian(2022)}]{gao2022learning}
Gao, X.; Zhang, Y.; and Tian, Y. 2022.
\newblock Learning to Incorporate Texture Saliency Adaptive Attention to Image Cartoonization.
\newblock In \emph{International Conference on Machine Learning}, 7183--7207.

\bibitem[{Ghosh and Chellappa(2016)}]{ghosh2016deep}
Ghosh, A.; and Chellappa, R. 2016.
\newblock Deep feature extraction in the DCT domain.
\newblock In \emph{International Conference on Pattern Recognition}, 3536--3541.

\bibitem[{Goodfellow et~al.(2014)Goodfellow, Pouget-Abadie, Mirza, Xu, Warde-Farley, Ozair, Courville, and Bengio}]{goodfellow2014generative}
Goodfellow, I.; Pouget-Abadie, J.; Mirza, M.; Xu, B.; Warde-Farley, D.; Ozair, S.; Courville, A.; and Bengio, Y. 2014.
\newblock Generative adversarial nets.
\newblock \emph{Advances in Neural Information Processing Systems}, 27.

\bibitem[{Hertz et~al.(2022)Hertz, Mokady, Tenenbaum, Aberman, Pritch, and Cohen-Or}]{hertz2022prompt}
Hertz, A.; Mokady, R.; Tenenbaum, J.; Aberman, K.; Pritch, Y.; and Cohen-Or, D. 2022.
\newblock Prompt-to-prompt image editing with cross attention control.
\newblock \emph{arXiv preprint arXiv:2208.01626}.

\bibitem[{Ho, Jain, and Abbeel(2020)}]{ho2020denoising}
Ho, J.; Jain, A.; and Abbeel, P. 2020.
\newblock Denoising diffusion probabilistic models.
\newblock \emph{Advances in Neural Information Processing Systems}, 33: 6840--6851.

\bibitem[{Isola et~al.(2017)Isola, Zhu, Zhou, and Efros}]{isola2017image}
Isola, P.; Zhu, J.-Y.; Zhou, T.; and Efros, A.~A. 2017.
\newblock Image-to-image translation with conditional adversarial networks.
\newblock In \emph{Proceedings of the IEEE Conference on Computer Vision and Pattern Recognition}, 1125--1134.

\bibitem[{Jiang et~al.(2019)Jiang, Lian, Tang, and Xiao}]{jiang2019scfont}
Jiang, Y.; Lian, Z.; Tang, Y.; and Xiao, J. 2019.
\newblock Scfont: Structure-guided Chinese font generation via deep stacked networks.
\newblock In \emph{Proceedings of the AAAI Conference on Artificial Intelligence}, 4015--4022.

\bibitem[{Karras, Laine, and Aila(2019)}]{karras2019style}
Karras, T.; Laine, S.; and Aila, T. 2019.
\newblock A style-based generator architecture for generative adversarial networks.
\newblock In \emph{Proceedings of the IEEE Conference on Computer Vision and Pattern Recognition}, 4401--4410.

\bibitem[{Kim, Kwon, and Ye(2022)}]{kim2022diffusionclip}
Kim, G.; Kwon, T.; and Ye, J.~C. 2022.
\newblock Diffusionclip: Text-guided diffusion models for robust image manipulation.
\newblock In \emph{Proceedings of the IEEE Conference on Computer Vision and Pattern Recognition}, 2426--2435.

\bibitem[{Kwon and Ye(2022)}]{kwon2022diffusion}
Kwon, G.; and Ye, J.~C. 2022.
\newblock Diffusion-based image translation using disentangled style and content representation.
\newblock \emph{arXiv preprint arXiv:2209.15264}.

\bibitem[{Lee et~al.(2020)Lee, Tseng, Mao, Huang, Lu, Singh, and Yang}]{lee2020drit++}
Lee, H.-Y.; Tseng, H.-Y.; Mao, Q.; Huang, J.-B.; Lu, Y.-D.; Singh, M.; and Yang, M.-H. 2020.
\newblock Drit++: Diverse image-to-image translation via disentangled representations.
\newblock \emph{International Journal of Computer Vision}, 128: 2402--2417.

\bibitem[{Lugmayr et~al.(2022)Lugmayr, Danelljan, Romero, Yu, Timofte, and Van~Gool}]{lugmayr2022repaint}
Lugmayr, A.; Danelljan, M.; Romero, A.; Yu, F.; Timofte, R.; and Van~Gool, L. 2022.
\newblock Repaint: Inpainting using denoising diffusion probabilistic models.
\newblock In \emph{Proceedings of the IEEE Conference on Computer Vision and Pattern Recognition}, 11461--11471.

\bibitem[{Luo and Hu(2021)}]{luo2021diffusion}
Luo, S.; and Hu, W. 2021.
\newblock Diffusion probabilistic models for 3d point cloud generation.
\newblock In \emph{Proceedings of the IEEE Conference on Computer Vision and Pattern Recognition}, 2837--2845.

\bibitem[{Mei and Patel(2023)}]{mei2023vidm}
Mei, K.; and Patel, V. 2023.
\newblock Vidm: Video implicit diffusion models.
\newblock In \emph{Proceedings of the AAAI Conference on Artificial Intelligence}, volume~37, 9117--9125.

\bibitem[{Meng et~al.(2021)Meng, He, Song, Song, Wu, Zhu, and Ermon}]{meng2021sdedit}
Meng, C.; He, Y.; Song, Y.; Song, J.; Wu, J.; Zhu, J.-Y.; and Ermon, S. 2021.
\newblock Sdedit: Guided image synthesis and editing with stochastic differential equations.
\newblock \emph{arXiv preprint arXiv:2108.01073}.

\bibitem[{Nichol et~al.(2022)Nichol, Dhariwal, Ramesh, Shyam, Mishkin, Mcgrew, Sutskever, and Chen}]{nichol2022glide}
Nichol, A.~Q.; Dhariwal, P.; Ramesh, A.; Shyam, P.; Mishkin, P.; Mcgrew, B.; Sutskever, I.; and Chen, M. 2022.
\newblock GLIDE: Towards Photorealistic Image Generation and Editing with Text-Guided Diffusion Models.
\newblock In \emph{International Conference on Machine Learning}, 16784--16804. PMLR.

\bibitem[{Park et~al.(2020)Park, Efros, Zhang, and Zhu}]{park2020contrastive}
Park, T.; Efros, A.~A.; Zhang, R.; and Zhu, J.-Y. 2020.
\newblock Contrastive learning for unpaired image-to-image translation.
\newblock In \emph{Proceedings of the European Conference on Computer Vision}, 319--345.

\bibitem[{Pizzati, Lalonde, and de~Charette(2022)}]{pizzati2022manifest}
Pizzati, F.; Lalonde, J.-F.; and de~Charette, R. 2022.
\newblock Manifest: Manifold deformation for few-shot image translation.
\newblock In \emph{Proceedings of the European Conference on Computer Vision}, 440--456.

\bibitem[{Radford et~al.(2021)Radford, Kim, Hallacy, Ramesh, Goh, Agarwal, Sastry, Askell, Mishkin, Clark et~al.}]{radford2021learning}
Radford, A.; Kim, J.~W.; Hallacy, C.; Ramesh, A.; Goh, G.; Agarwal, S.; Sastry, G.; Askell, A.; Mishkin, P.; Clark, J.; et~al. 2021.
\newblock Learning transferable visual models from natural language supervision.
\newblock In \emph{International Conference on Machine Learning}, 8748--8763.

\bibitem[{Ramesh et~al.(2022)Ramesh, Dhariwal, Nichol, Chu, and Chen}]{ramesh2022hierarchical}
Ramesh, A.; Dhariwal, P.; Nichol, A.; Chu, C.; and Chen, M. 2022.
\newblock Hierarchical text-conditional image generation with clip latents.
\newblock \emph{arXiv preprint arXiv:2204.06125}, 1(2): 3.

\bibitem[{Richardson et~al.(2021)Richardson, Alaluf, Patashnik, Nitzan, Azar, Shapiro, and Cohen-Or}]{richardson2021encoding}
Richardson, E.; Alaluf, Y.; Patashnik, O.; Nitzan, Y.; Azar, Y.; Shapiro, S.; and Cohen-Or, D. 2021.
\newblock Encoding in style: a stylegan encoder for image-to-image translation.
\newblock In \emph{Proceedings of the IEEE Conference on Computer Vision and Pattern Recognition}, 2287--2296.

\bibitem[{Roich et~al.(2022)Roich, Mokady, Bermano, and Cohen-Or}]{roich2022pivotal}
Roich, D.; Mokady, R.; Bermano, A.~H.; and Cohen-Or, D. 2022.
\newblock Pivotal tuning for latent-based editing of real images.
\newblock \emph{ACM Transactions on Graphics}, 42: 1--13.

\bibitem[{Rombach et~al.(2022)Rombach, Blattmann, Lorenz, Esser, and Ommer}]{rombach2022high}
Rombach, R.; Blattmann, A.; Lorenz, D.; Esser, P.; and Ommer, B. 2022.
\newblock High-resolution image synthesis with latent diffusion models.
\newblock In \emph{Proceedings of the IEEE Conference on Computer Vision and Pattern Recognition}, 10684--10695.

\bibitem[{Saharia et~al.(2022{\natexlab{a}})Saharia, Chan, Chang, Lee, Ho, Salimans, Fleet, and Norouzi}]{saharia2022palette}
Saharia, C.; Chan, W.; Chang, H.; Lee, C.; Ho, J.; Salimans, T.; Fleet, D.; and Norouzi, M. 2022{\natexlab{a}}.
\newblock Palette: Image-to-image diffusion models.
\newblock In \emph{ACM SIGGRAPH Conference Proceedings}, 1--10.

\bibitem[{Saharia et~al.(2022{\natexlab{b}})Saharia, Chan, Saxena, Li, Whang, Denton, Ghasemipour, Gontijo~Lopes, Karagol~Ayan, Salimans et~al.}]{saharia2022photorealistic}
Saharia, C.; Chan, W.; Saxena, S.; Li, L.; Whang, J.; Denton, E.~L.; Ghasemipour, K.; Gontijo~Lopes, R.; Karagol~Ayan, B.; Salimans, T.; et~al. 2022{\natexlab{b}}.
\newblock Photorealistic text-to-image diffusion models with deep language understanding.
\newblock \emph{Advances in neural information processing systems}, 35: 36479--36494.

\bibitem[{Saharia et~al.(2022{\natexlab{c}})Saharia, Ho, Chan, Salimans, Fleet, and Norouzi}]{saharia2022image}
Saharia, C.; Ho, J.; Chan, W.; Salimans, T.; Fleet, D.~J.; and Norouzi, M. 2022{\natexlab{c}}.
\newblock Image super-resolution via iterative refinement.
\newblock \emph{IEEE Transactions on Pattern Analysis and Machine Intelligence}, 45: 4713--4726.

\bibitem[{Song, Meng, and Ermon(2020)}]{song2020denoising}
Song, J.; Meng, C.; and Ermon, S. 2020.
\newblock Denoising diffusion implicit models.
\newblock \emph{arXiv preprint arXiv:2010.02502}.

\bibitem[{Tumanyan et~al.(2022)Tumanyan, Bar-Tal, Bagon, and Dekel}]{tumanyan2022splicing}
Tumanyan, N.; Bar-Tal, O.; Bagon, S.; and Dekel, T. 2022.
\newblock Splicing vit features for semantic appearance transfer.
\newblock In \emph{Proceedings of the IEEE Conference on Computer Vision and Pattern Recognition}, 10748--10757.

\bibitem[{Tumanyan et~al.(2023)Tumanyan, Geyer, Bagon, and Dekel}]{tumanyan2023plug}
Tumanyan, N.; Geyer, M.; Bagon, S.; and Dekel, T. 2023.
\newblock Plug-and-play diffusion features for text-driven image-to-image translation.
\newblock In \emph{Proceedings of the IEEE Conference on Computer Vision and Pattern Recognition}, 1921--1930.

\bibitem[{Wang et~al.(2022)Wang, Zhang, Zhang, Ouyang, Chen, Chen, and Wen}]{wang2022pretraining}
Wang, T.; Zhang, T.; Zhang, B.; Ouyang, H.; Chen, D.; Chen, Q.; and Wen, F. 2022.
\newblock Pretraining is all you need for image-to-image translation.
\newblock \emph{arXiv preprint arXiv:2205.12952}.

\bibitem[{Wu, Lischinski, and Shechtman(2021)}]{wu2021stylespace}
Wu, Z.; Lischinski, D.; and Shechtman, E. 2021.
\newblock Stylespace analysis: Disentangled controls for stylegan image generation.
\newblock In \emph{Proceedings of the IEEE Conference on Computer Vision and Pattern Recognition}, 12863--12872.

\bibitem[{Xie et~al.(2021)Xie, Song, Xu, Xu, Zhang, and Wang}]{xie2021learning}
Xie, W.; Song, D.; Xu, C.; Xu, C.; Zhang, H.; and Wang, Y. 2021.
\newblock Learning frequency-aware dynamic network for efficient super-resolution.
\newblock In \emph{Proceedings of the IEEE International Conference on Computer Vision}, 4308--4317.

\bibitem[{Yi et~al.(2019)Yi, Liu, Lai, and Rosin}]{yi2019apdrawinggan}
Yi, R.; Liu, Y.-J.; Lai, Y.-K.; and Rosin, P.~L. 2019.
\newblock Apdrawinggan: Generating artistic portrait drawings from face photos with hierarchical gans.
\newblock In \emph{Proceedings of the IEEE Conference on Computer Vision and Pattern Recognition}, 10743--10752.

\bibitem[{Zhang and Agrawala(2023)}]{zhang2023adding}
Zhang, L.; and Agrawala, M. 2023.
\newblock Adding conditional control to text-to-image diffusion models.
\newblock \emph{arXiv preprint arXiv:2302.05543}.

\bibitem[{Zhang et~al.(2022)Zhang, Chen, Tu, Wan, Xu, and Ma}]{zhang2022wavelet}
Zhang, L.; Chen, X.; Tu, X.; Wan, P.; Xu, N.; and Ma, K. 2022.
\newblock Wavelet knowledge distillation: Towards efficient image-to-image translation.
\newblock In \emph{Proceedings of the IEEE Conference on Computer Vision and Pattern Recognition}, 12464--12474.

\bibitem[{Zhu et~al.(2017)Zhu, Park, Isola, and Efros}]{zhu2017unpaired}
Zhu, J.-Y.; Park, T.; Isola, P.; and Efros, A.~A. 2017.
\newblock Unpaired image-to-image translation using cycle-consistent adversarial networks.
\newblock In \emph{Proceedings of the IEEE International Conference on Computer Vision}, 2223--2232.

\end{thebibliography}

\end{document}